%% file: template.tex
\documentclass{article}

\usepackage{arxiv}

\usepackage[utf8]{inputenc}
\usepackage[T1]{fontenc}
\usepackage{url}
\usepackage{booktabs}
\usepackage{amsfonts}
\usepackage{amsmath,amsfonts}
\usepackage{mathtools}
\usepackage{nicefrac}
\usepackage{microtype}
\usepackage{lipsum}
\usepackage{graphicx}  
\usepackage{natbib}
\usepackage{doi}
\usepackage{algorithm}
\usepackage{algorithmicx}
\usepackage{algpseudocode}
\usepackage{subfig}
\usepackage{xcolor}
\usepackage{array}
\usepackage{multirow}
\usepackage[inline]{enumitem}
\usepackage{rotating}
\usepackage{caption}
\usepackage{tikz}
\usepackage{tipa}
\usepackage[table]{xcolor}
\usepackage{svg}
\usepackage{hhline}
\usepackage{csquotes}

\usepackage{hyperref}
\usepackage{cleveref}
\usepackage[nohyperlinks, nolist]{acronym}

\usepackage{amsthm}
\theoremstyle{definition}
\newtheorem{definition}{Definition}

\title{No Single Metric Tells the Whole Story: A Multi-Dimensional Evaluation Framework for Uncertainty Attributions}

\date{} 					

\author{Emily Schiller \\
	XITASO GmbH IT \& Software Solutions, \\ The Artificial Intelligence and Cognitive Load Research Lab \\ University College Cork \\ Augsburg, Germany \\
	\texttt{emily.schiller@xitaso.com} \\
	\And
    Teodor Chiaburu \\
	Berliner Hochschule für Technik \\ 
    Berlin, Germany\\
	\texttt{chiaburu.teodor@bht-berlin.de} \\
    \And
	Marco Zullich \\
	Faculty of Technology, Policy and Management \\
    Delft University of Technology\\
    Delft, the Netherlands \\
	\texttt{m.zullich@tudelft.nl} \\
    \And
    Luca Longo \\
	The Artificial Intelligence and Cognitive Load Research Lab, \\ 
    School of Computer Science and Information Technology\\
    University College Cork \\
    Cork, Ireland
	\texttt{llongo@ucc.ie} \\
}

\begin{document}
\maketitle
\begin{abstract}
Research on explainable AI (XAI) has frequently focused on explaining model predictions. More recently, methods have been proposed to explain prediction uncertainty by attributing it to input features (uncertainty attributions). However, the evaluation of these methods remains inconsistent as studies rely on heterogeneous proxy tasks and metrics, hindering comparability. We address this by aligning uncertainty attributions with the well-established Co-12 framework for XAI evaluation. We propose concrete implementations for the correctness, consistency, continuity, and compactness properties. Additionally, we introduce \emph{conveyance}, a property tailored to uncertainty attributions that evaluates whether controlled increases in epistemic uncertainty reliably propagate to feature-level attributions. We demonstrate our evaluation framework with eight metrics across combinations of uncertainty quantification and feature attribution methods on tabular and image data. Our experiments show that gradient-based methods consistently outperform perturbation-based approaches in consistency and conveyance, while Monte-Carlo dropconnect outperforms Monte-Carlo dropout in most metrics. Although most metrics rank the methods consistently across samples, inter-method agreement remains low. This suggests no single metric sufficiently evaluates uncertainty attribution quality. The proposed evaluation framework contributes to the body of knowledge by establishing a foundation for systematic comparison and development of uncertainty attribution methods.
\end{abstract}


\begin{acronym}
\acro{ai}[AI]{artificial intelligence}
\acro{cnn}[CNN]{convolutional neural network}
\acro{dnn}[DNN]{deep neural network}
\acroplural{dnn}[DNNs]{deep neural networks}
\acro{mlp}[MLP]{multi-layer perceptron}
\acro{rnn}[RNN]{recurrent neural network}
\acro{xai}[XAI]{Explainable Artificial Intelligence}
\acro{xuq}[XUQ]{Explainable Uncertainty Quantification}
\acro{uq}[UQ]{Uncertainty Quantification}
\acro{mse}[MSE]{Mean Squared Error}
\acro{mae}[MAE]{Mean Absolute Error}
\acro{rmse}[RMSE]{Root Mean Squared Error}
\acro{ood}[OOD]{out-of-distribution}
\acro{auc}[AUC]{area under the curve}
\acro{rri}[RRI]{Relative Rank Improvement}
\acro{ucs}[UCS]{Uncertainty Conveyance Similarity}
\acro{mcd}[MCD]{Monte-Carlo dropout}
\acro{mcdc}[MCDC]{Monte-Carlo dropconnect}

\end{acronym}

\input{introduction}

\input{related_work}

\input{background}

\input{method}

\input{experiments}

\input{results_new}

\input{discussion_new}

\input{conclusion}

\subsubsection*{Acknowledgements} E. Schiller acknowledges funding support from the Bavarian State Ministry of Economic Affairs, Regional Development and Energy under grant number 41-6618c/570/2-LSM-2203-0010.

\bibliographystyle{unsrt}
\bibliography{references}  






\end{document}

%% file: introduction.tex
\section{Introduction}
The field of \ac{xai} has grown into a highly active area of research. A substantial body of \ac{xai} research focuses on explaining why a model produces a particular prediction~\cite{clue}. In parallel, the field of \ac{uq} aims at quantifying the uncertainties inherent in \ac{ai} predictions. UQ methods typically consider three types of uncertainty: epistemic, aleatoric, and predictive uncertainty. Epistemic uncertainty (\emph{model uncertainty}) arises from limited knowledge or insufficient data and can be reduced by collecting more data or improving the model; aleatoric uncertainty (\emph{data uncertainty}) arises from the data-generating process itself and cannot be reduced; predictive uncertainty combines both sources~\cite{hullermeier_aleatoric_epistemic}. \ac{xai} and \ac{uq} are fields dedicated to enhancing the reliability of \ac{ai} models, promoting responsible
\ac{ai} usage, and, in turn, aiming to achieve user trust. Additionally, these fields enable developers to evaluate their models, uncover weaknesses, and ultimately improve performance. The intersection of \ac{xai} and \ac{uq} holds significant potential to further advance these objectives.
 
Recent research has begun to explore this intersection by proposing methods that combine \ac{xai} and \ac{uq} techniques~\cite{bley2025explaining,iversen2023identifying,wang_semantic_2023}. The emerging research field of \ac{xuq} focuses on explaining uncertainty in model predictions rather than the predictions themselves. Consider, for example, a medical diagnosis system. Such a system should indicate which input patterns lead to high uncertainty, enabling doctors to make informed decisions and \enquote{advise the clinicians on the course of action needed to improve the AI predictions}~\cite{EUAI4healthcare}. Most prior work in \ac{xuq} has focused on explaining uncertainty in terms of input features, an approach we denote as uncertainty attributions~\cite{brown_using_2022,wang_semantic_2023,iversen2023identifying,wang_gradient-based_2023,watson_explaining_2023,bley2025explaining}. Similar to feature attributions in \ac{xai}, these methods attribute importance scores to input features. However, the importance reflects the contribution to prediction uncertainty rather than to the prediction itself.

Despite growing interest in \ac{xuq}, the evaluation of uncertainty attribution methods remains fragmented and inconsistent. While the \ac{xai} literature has converged towards established evaluation protocols~\cite{nauta2023anecdotal}, \ac{xuq} research currently relies on heterogeneous proxy tasks and metrics that vary across studies~\cite{bley2025explaining,iversen2023identifying,wang_gradient-based_2023}. This lack of standardisation creates several problems. First, it hinders systematic comparison of methods. Second, it limits reproducibility. Third, it complicates the assessment of progress in the field. The challenges are further compounded by underlying questions about what defines a \enquote{good} uncertainty attribution.

The \ac{xai} community has made significant progress in addressing analogous challenges in explaining model predictions. The field has distinguished among functionally grounded metrics, which evaluate explanation quality based on model behaviour; human-grounded evaluation, based on user studies and simplified tasks; and application-grounded evaluation, involving users and a real-world application task~\cite{doshi-velez_towards_2018}. Functionally grounded metrics enable objective, reproducible evaluation and have become standard in \ac{xai} research for initial technical inspection of \ac{xai} methods~\cite{nauta2023anecdotal}. Nauta et al.~\cite{nauta2023anecdotal} performed an exhaustive literature review on functionally grounded metrics, which they synthesised into the Co-12 framework, including twelve desirable properties of explanations. 

In this work, we investigate the transferability of functionally grounded \ac{xai} evaluation properties and metrics to uncertainty attributions. We adapt established \ac{xai} metrics, aligned with the Co-12 properties, to the setting of uncertainty attribution. Specifically, we propose concrete implementations for four properties: correctness, consistency, continuity, and compactness (see~\Cref{section:properties}). We omit the other eight because their implementation depends on the specific application or user involvement, or because they pose conceptual challenges.
Recognising that uncertainty attributions require additional considerations, we introduce a new property, \emph{conveyance}, that assesses whether controlled increases in uncertainty reliably propagate to feature-level uncertainty attributions. We demonstrate our evaluation framework by comparing multiple uncertainty attribution methods across two datasets (UCI Wine Quality\footnote{Available at https://archive.ics.uci.edu/dataset/186/wine+quality}, MNIST\footnote{Available at http://yann.lecun.com/exdb/mnist}). We further validate the framework by applying sanity checks to assess the reliability of the proposed metrics. The contributions of this work are as follows: 
\begin{itemize}
    \item We provide a systematic adaptation of functionally grounded \ac{xai} evaluation metrics for uncertainty attribution. This includes proposing concrete metrics for the four Co-12 properties and introducing the novel \emph{conveyance} property.
    \item We introduce a novel metric, \emph{uncertainty similarity conveyance}, to evaluate the conveyance property for uncertainty.
    \item We conduct a systematic empirical evaluation demonstrating how different uncertainty attribution methods perform across multiple quality dimensions and apply sanity checks to assess metric reliability.
\end{itemize}

The remainder of this paper is organised as follows. \Cref{section:relatedwork} reviews related work. \Cref{section:preliminaries} introduces formal notions and the used uncertainty attribution approach. \Cref{section:properties} presents our evaluation framework with adapted evaluation properties and corresponding metrics. Section~\ref{section:experiments} describes the experimental setup. \Cref{section:results} and~\ref{section:discussion} report and discuss the results and limitations. Finally,~\Cref{section:conclusion} concludes with directions for future work. The code for the experiments is available at \url{https://github.com/emilyS135/xuq_eval}.

%% file: related_work.tex
\section{Related Work}
\label{section:relatedwork}
Previous work on \ac{xuq} has primarily focused on developing novel methods to generate explanations of prediction uncertainty~\cite {clue,watson_explaining_2023,bley2025explaining}. One of the first approaches to explain predictive uncertainty was CLUE by Antorán et al.~\cite{clue}, a counterfactual method for differentiable probabilistic models like Bayesian Neural Networks (BNNs). It modifies inputs based on uncertain predictions using a generative model, with attribution maps computed by measuring differences between original and modified inputs. The predominant approach in \ac{xuq}, however, is uncertainty attribution, which adapts feature attribution methods from traditional \ac{xai} to explain uncertainty estimates~\cite{brown_using_2022,patro_u-cam_2019,wang_semantic_2023,iversen2023identifying,wang_gradient-based_2023,watson_explaining_2023,bley2025explaining,perez_attribution_2022,explainable-competency}.

\subsection{Explaining Uncertainty with Uncertainty Attributions}

Uncertainty attribution methods have been applied across various uncertainty types and model architectures. Depending on the underlying \ac{uq} approach, these methods can explain either predictive, epistemic, or aleatoric uncertainty by identifying which input features contribute most to the model's uncertainty. UA-Backprop by Wang et al.~\cite{wang_gradient-based_2023} leverages gradient information to produce uncertainty attributions for Bayesian deep learning models, accounting for both epistemic and aleatoric uncertainty. The evaluation employs perturbation-based metrics: pixels are revealed or blurred in attribution order while tracking changes in uncertainty. Watson et al.~\cite{watson_explaining_2023} investigated the application of Shapley Values to explain aleatoric and epistemic uncertainty. Their evaluation includes a visual inspection of the uncertainty attributions and a downstream task demonstrating the utility of active feature selection. Additionally, they assess the method's ability to detect covariate shifts by perturbing features with Gaussian noise to alter their distributions and measuring changes in uncertainty attribution relative to the original sample.

Bley et al.~\cite{bley2025explaining} introduced an ensemble-based framework that generates feature attributions for each ensemble member and calculates their covariance matrix to explain predictive uncertainty. Their evaluation uses feature flipping, iteratively setting high-attribution features to baseline values while tracking changes in uncertainty. The authors further demonstrate practical utility by identifying underrepresented features in a model at test time and showing how retraining on a consolidated dataset reduces uncertainty.
Iversen et al.~\cite{iversen2023identifying} present a method specifically for explaining aleatoric uncertainty by adding a variance output neuron to pre-trained models and applying feature attribution methods to this variance output. They assess their method based on three desiderata: robustness, faithfulness, and localisation. To measure the method's ability to highlight relevant features, they use synthetic and semi-synthetic datasets perturbed with known noise. Robustness is measured using local Lipschitz continuity. To assess faithfulness, they evaluate whether perturbing relevant features results in a significant reduction in predictive performance.

In summary, evaluation practices for uncertainty attributions vary substantially across studies, with most studies focusing on faithfulness metrics that track changes in uncertainty after perturbing high-attribute features. We will describe and categorise existing evaluation metrics for \ac{xuq} in more detail in \Cref{section:properties}.

\subsection{Quantitative Evaluation of Feature Attributions}

Similar to \ac{xuq}, the evaluation of \ac{xai} methods has been inconsistent across the literature, which hinders the comparability of different approaches~\cite{longo_explainable_2024,nauta2023anecdotal,alangari_survey}. The literature shows no consensus on the most important explanation desiderata. Many authors advocate customising explanation properties to specific use cases and stakeholder needs~\cite{longo_explainable_2024,poursabzi-sangdehManipulatingMeasuringModel2018}. Nevertheless, faithfulness is frequently identified as a critical property. Faithfulness metrics measure how well an explanation captures the model's actual behaviour. A high-fidelity method assigns high relevance to features that significantly reduce the model's confidence when removed. Conversely, it assigns low relevance to features that have minimal impact when removed. However, prior work has investigated the trustworthiness of faithfulness metrics~\cite{tomsett_sanity_2020}. Tomsett et al.~\cite{tomsett_sanity_2020} show that these metrics can be statistically unreliable and inconsistent. Therefore, faithfulness should not be used as the sole criterion for evaluation.
Several comprehensive surveys and benchmarks have been proposed to address these challenges by suggesting multiple desiderata for explanations~\cite{doshi-velez_towards_2018,ali_explainable_2023,nauta2023anecdotal}. A particularly influential contribution to standardising \ac{xai} evaluation is the work by Nauta et al.~\cite{nauta2023anecdotal}, who conducted an exhaustive survey of papers that use quantitative metrics to evaluate \ac{xai} methods. They categorised existing metrics according to 12 conceptual properties (``Co-12''). These represent desiderata that explanations should fulfil, such as correctness, continuity and compactness. The Co-12 aims to guide and standardise the evaluation of \ac{xai}. Several software frameworks have been developed to facilitate the practical evaluation of \ac{xai} methods. Quantus~\cite{hedstrom_quantus_2023} is among the most comprehensive frameworks, offering over 30 evaluation metrics. Other \ac{xai} evaluation toolkits include Ablation~\cite{hameed_ablation} and OpenXAI~\cite{agarwal2022openxai}. A comparative survey of these toolkits assessed their coverage of the Co-12 properties, revealing that Quantus achieved the highest coverage with 6 out of 12 properties~\cite{le_benchmarking_2023}.

More recently, Monke et al.~\cite{monke2025confusion} presented a framework specifically designed for evaluating prototype-based \ac{xai} methods. They discussed the applicability of Co-12 properties to prototype-based explanations, proposed additional properties tailored to this explanation paradigm, and introduced corresponding metrics to quantify these properties. Our evaluation framework follows a similar approach: we systematically consider the relevance and applicability of established properties and metrics from the \ac{xai} evaluation literature, particularly the Co-12 properties, in the context of uncertainty attributions. Building on this foundation, our aim is to design an evaluation framework for \ac{xuq} methods that accounts for the similarities and unique characteristics of uncertainty explanations compared to traditional feature attributions.

%% file: background.tex
\section{Formal Notions}
\label{section:preliminaries}

In the following, we present the key concepts that underpin our work. Given a classification or regression task and a dataset $\mathcal{D} = \{(x_i,y_i)\}_{i=1}^m$ with $x_i\in \mathcal{X} \subseteq \mathbb{R}^n$ being the input samples and $y_i \in \mathcal{Y} \subseteq \mathbb{R}$ the labels, we define a model with ensemble-based \ac{uq} as $f : \mathcal{X} \to \mathcal{Y}^K$,  
where $K$ defines the ensemble size. For each input sample $x$, the model $f$ outputs $K$ predictions on the same space $\mathcal{Y}$. To obtain a final prediction, we average these $K$ predictions. For regression, we compute the mean of all $K$ ensemble outputs.
For classification, we average the softmax probabilities $\hat{p}^{(k)}$, $k=1,...,K$, and select the class with the highest value in the averaged probability distribution, $\hat{y}  = \arg\max_{y\in \mathcal{Y}}\frac{1}{K} \sum_{k=1}^K{\hat{p}^{(k)}_y}.$ To obtain an estimate of predictive uncertainty, we use the regression predictive variance and the variance of the softmax probabilities for the predicted class in classification, both denoted as $s^2$. The $K$ predictions can be obtained, for example, from deep ensembles~\cite{lakshminarayanan2017simple}, \ac{mcd}~\cite{gal16dropout}, or \ac{mcdc}~\cite{mobiny2021dropconnect}. Due to the high computational cost of deep ensembles, we focus on \ac{mcd} and \ac{mcdc}.

\begin{figure}[t]
    \centering    
    \includegraphics[width=\columnwidth]{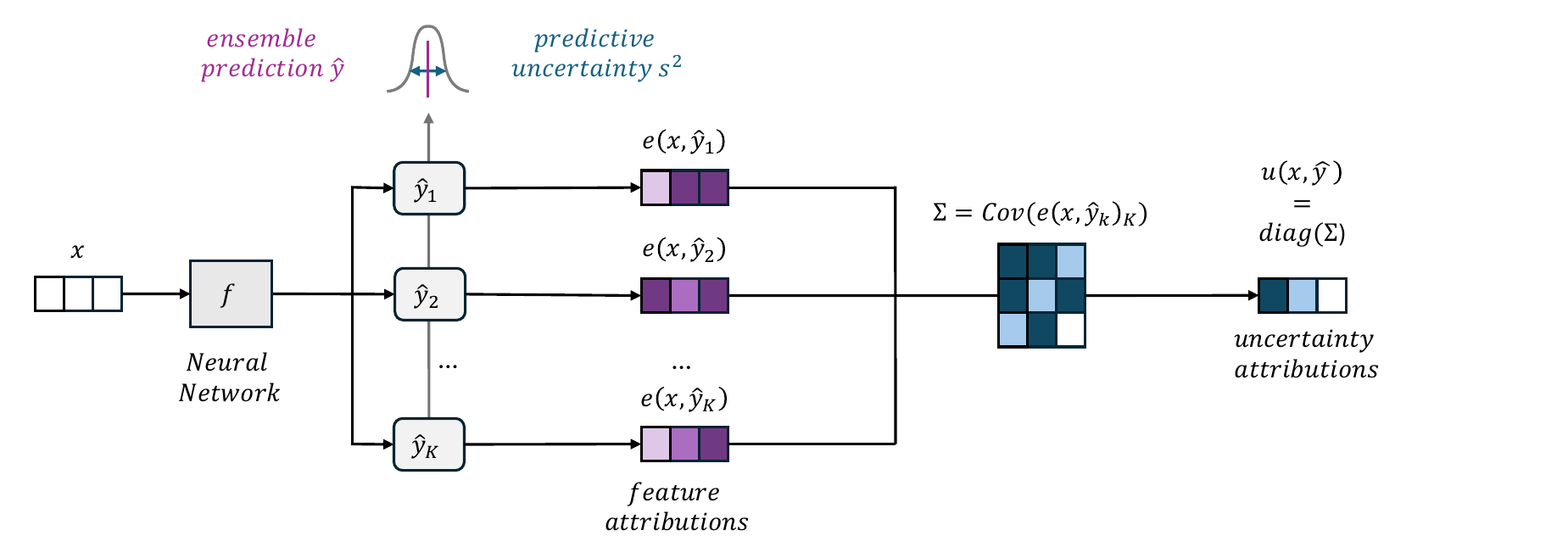}
    \caption{Uncertainty attribution method by Bley et al.~\cite{bley2025explaining}. Predictive variance is estimated via ensemble-based UQ. Each of $K$ predictions is explained, and uncertainty attributions are derived as diagonal elements of the covariance matrix of these explanations. Illustration adapted from Bley et al.~\cite{bley2025explaining}.}
    \label{fig:uncertainty_attribution_method}
\end{figure}

\begin{figure}[tb]
    \centering    \includegraphics[width=\columnwidth]{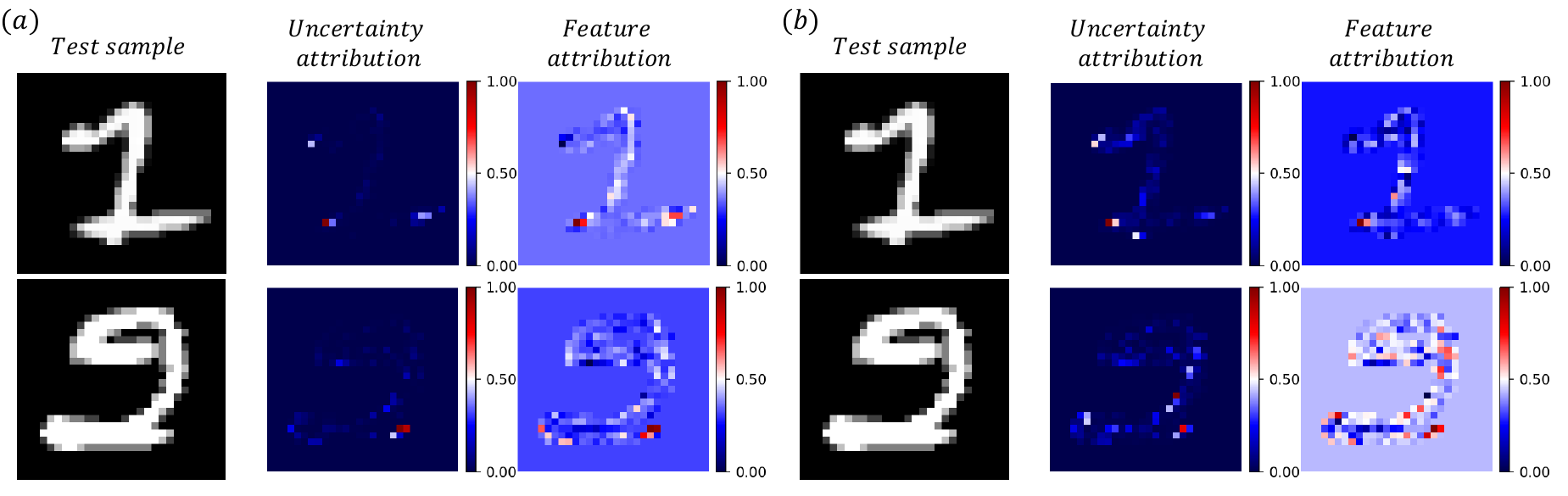}
    \caption{Uncertainty and feature attributions using \ac{mcd} and (a) LRP and (b) Integrated Gradients for test samples from the MNIST dataset. Dark blue pixels indicate low importance and dark red pixels indicate high importance to predictive uncertainty (uncertainty attribution) or the prediction (feature attribution).}
    \label{fig:example_attributions}
\end{figure}

We follow the framework proposed by Bley et al.~\cite{bley2025explaining} to generate uncertainty attributions (see~\Cref{fig:uncertainty_attribution_method}): consider an input $x$ and a fixed feature attribution method (e.g., Integrated Gradients~\cite{sundararajan2017axiomatic}). Let $e(x, \hat{y}_k)_k \in \mathbb{R}^n$ denote the feature attribution vector for the $k$-th ensemble member's prediction $\hat{y}_k$. Given the covariance matrix of these $K$ feature attributions $\Sigma\in \mathbb{R}^{n\times n}$, we define the \emph{uncertainty attribution} of $x$ as the diagonal of $\Sigma$, $u(x, \hat{y}) \coloneqq \text{diag}(\Sigma)$. Thus, an uncertainty attribution method consists of an ensemble-based \ac{uq} method and a feature attribution method. Note that while variance of explanations is typically viewed as \textit{uncertainty in the explanation}\cite{mulye2025uncertainty,chiaburu2025uncertainty}, the authors show that these concepts can be equated under specific constraints. The validity of this approach (Equation 5 in~\cite{bley2025explaining}) requires the feature attribution method to be linear with respect to the prediction $\hat{y}$ and obey conservation properties. Thus, it holds for deterministic methods (e.g., LRP~\cite{bach2015pixel} or Integrated Gradients~\cite{sundararajan2017axiomatic}) but fails for stochastic or non-linear explainers like LIME~\cite{ribeiro_why_2016} or SHAP~\cite{lundberg_unified_2017} implementations based on stochastic sampling, as discussed in~\Cref{section:discussion}. The method is restricted to ensemble-based \ac{uq} using variance of predictions as the uncertainty estimate. \Cref{fig:example_attributions} shows feature and uncertainty attributions for selected MNIST samples. The colour scale encodes attribution values, where red indicates high importance and blue indicates low importance of a pixel for the model's prediction (feature attributions) or predictive uncertainty (uncertainty attributions).

%% file: method.tex
\section{Design of an Evaluation Framework for Uncertainty Attributions}
\label{section:properties}

\begin{figure}[b!]
    \centering    \includegraphics[width=\columnwidth]{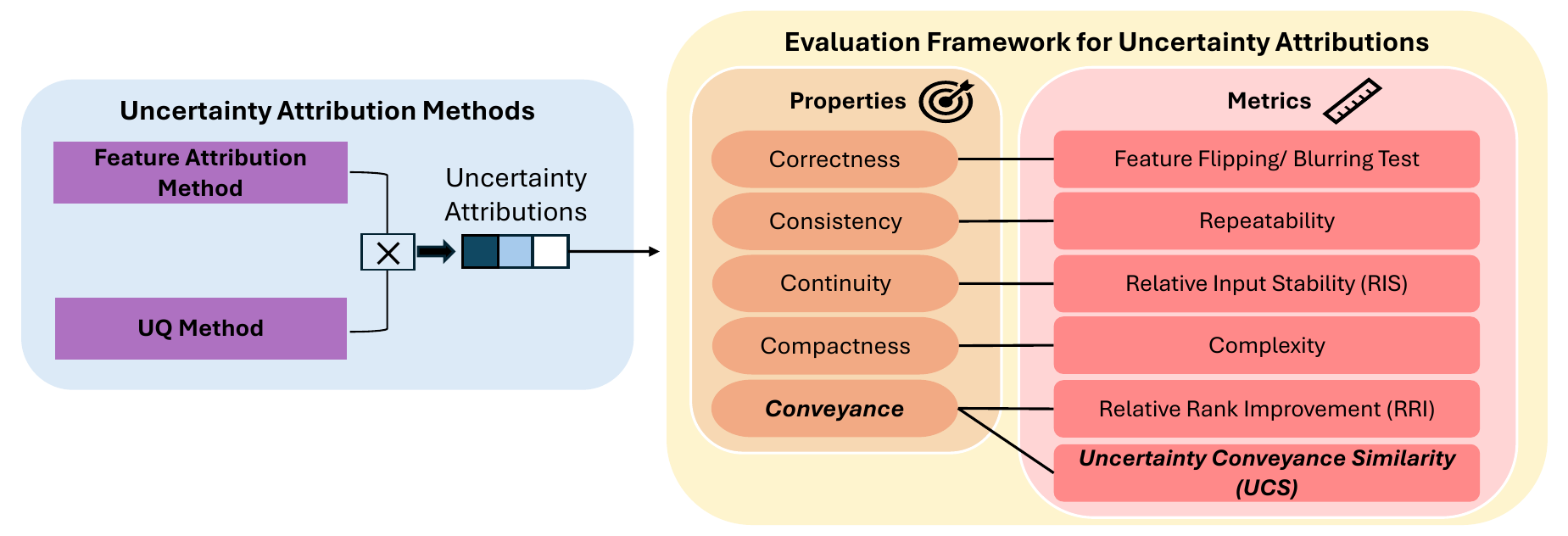}
    \caption{Overview of the proposed evaluation framework. We combine a feature attribution method with an \ac{uq} method (blue panel) to generate uncertainty attributions. We evaluate these uncertainty attributions by adapting and extending properties (orange panel) and metrics (red panel) from feature attribution to the uncertainty setting; novel categories and metrics appear in italics. Metric results are aggregated into an overall score for each uncertainty attribution method.}
    \label{fig:evaluation_framework}
\end{figure}

Our proposed evaluation framework (see~\Cref{fig:evaluation_framework}) uses the Co-12 properties of Nauta et al.~\cite{nauta2023anecdotal} as its foundational structure. We focus specifically on uncertainty attributions as they represent the most prominent class of explanations in the reviewed literature on \ac{xuq}. However, a key challenge is that Co-12 properties were formulated to explain model predictions (including feature attributions), whereas uncertainty attributions explain the prediction uncertainty. Consequently, the applicability of Co-12 properties to uncertainty attribution is not straightforward and may require reinterpretation and adaptation.
For each of these properties, we 
\begin{enumerate*}[label=(\roman*)]
    \item define the property in the context of uncertainty attributions, 
    \item review how it has been evaluated in related work, and 
    \item propose concrete metrics drawn from prior literature on uncertainty attribution or feature attribution. 
\end{enumerate*}
 Additionally, we introduce a \ac{xuq}-specific property, \emph{conveyance}, that is not adequately covered by the existing Co-12 taxonomy, and we suggest corresponding evaluation metrics. \Cref{tab:used_properties} provides a summary of all operationalised Co-12 properties and their associated metrics.
The remaining properties are not considered in our evaluation framework because they require user involvement or pose conceptual challenges when applied to uncertainty attribution and thus fall outside the scope of this analysis. Addressing these properties represents an important avenue for future work.

\begin{table}[t]
\centering
\caption{Evaluation properties and metrics for uncertainty attributions. We adapt four Co-12 properties~\cite{nauta2023anecdotal} for evaluating uncertainty attribution methods. For each property, we report metrics from the prior uncertainty-attribution literature and from our framework. Conveyance (highlighted in blue) is our proposed property specific to uncertainty attributions. Bold metrics are novel adaptations from feature attribution research to uncertainty attributions. Bold italics indicate our novel uncertainty similarity conveyance metric.}
\setlength{\tabcolsep}{8pt} 
\begin{tabular}{p{1.7cm} p{6.4cm} p{6cm}}
\hline
\textbf{Co-12 \newline Property} &  \textbf{Metrics used in Literature on uncertainty attributions} & \textbf{Metrics selected for our Evaluation Framework} \\

\hline
correctness \newline\ref{prop:correctness}  &
feature flipping~\cite{bley2025explaining}, \newline blurring test~\cite{perez_attribution_2022,wang_gradient-based_2023,iversen2023identifying} &
feature flipping\\
\hline
consistency \newline\ref{prop:consistency}  &
\multicolumn{1}{c}{-}  &
\textbf{repeatability} \\
\hline
continuity \newline\ref{prop:continuity}  &
 local Lipschitz continuity~\cite{iversen2023identifying} &
\textbf{relative input stability (RIS)}\\
\hline
compactness \newline\ref{prop:compactness} &
\multicolumn{1}{c}{-}  &
\textbf{complexity} \\
\hline
\rowcolor{blue!25}
conveyance \newline\ref{prop:conveyance}& distribution shift detection \newline\cite{brown_using_2022,wang_semantic_2023,watson_explaining_2023,explainable-competency}, \newline controlled aleatoric uncertainty \cite{iversen2023identifying} &
\textbf{relative rank improvement (RRI)}, \newline \textbf{\textit{uncertainty conveyance similarity (UCS) }}\\
\hline

\end{tabular}

\label{tab:used_properties}
\end{table}

\subsection{Correctness}
\label{prop:correctness}
Correctness assesses how faithfully an uncertainty attribution reflects the model's mechanism for estimating predictive uncertainty. This property is also commonly known as faithfulness or fidelity \cite{nauta2023anecdotal,tomsett_sanity_2020}. Since uncertainty attributions inform decisions such as data collection, human review, or prediction trust~\cite{wang_gradient-based_2023,bley2025explaining}, unfaithful attributions can mislead these actions. Prior work has adapted faithfulness metrics from feature attribution evaluation to uncertainty attributions, including feature flipping for tabular and blurring tests for image data~\cite{bley2025explaining,wang_gradient-based_2023,perez_attribution_2022}. 

In our framework, we apply a deletion-style faithfulness metric, which we refer to as feature flipping. We operationalize this as follows:

\begin{enumerate}
\item For a test sample $x$, we compute its predictive uncertainty $s^2$ and uncertainty attribution $u(x,\hat{y})$, then sort features by descending attribution.
\item We iteratively replace the top-ranked features with a baseline and track the change in predictive uncertainty $s^2$ after each removal. We track the \ac{auc} averaged over all considered test samples. For tabular data, we use conditional resampling based on the training data~\cite{bley2025explaining}; for images, we use Gaussian blur to remove information while maintaining spatial structure~\cite{perez_attribution_2022}.
\item We compute the curve of uncertainty change vs. number of removed features and calculate the area under the curve (AUC), averaged over all test samples. Lower AUC indicates more faithful attributions, meaning that highly-attributed features reduce uncertainty.
\end{enumerate}


\subsection{Consistency}
\label{prop:consistency}

The key idea with consistency is that identical inputs should have identical explanations, i.e., that explanations are deterministic and implementation-invariant. The latter means that two models that generate the same output for the same input should provide the same explanations, regardless of their architectures~\cite{nauta2023anecdotal}. In XUQ, non-determinism can arise from multiple sources. First, using Bayesian \ac{uq} methods, such as \ac{mcd} or \ac{mcdc}, will introduce stochasticity. Additionally, the explanation method itself can introduce stochasticity; for example, LIME~\cite{ribeiro_why_2016} relies on stochastic sampling and surrogate model training~\cite{chiaburu2025uncertainty}. To test to what extent uncertainty attributions deviate, we compute them $M>0$ times on the same input, but with different seeds, and compute their similarity:
\begin{align}
    \text{repeatability}(x, u) = \frac{1}{M} \sum_{m=1}^M\text{sim}(u, u^m) ,
\end{align}
where $\textnormal{sim}(\cdot,\cdot)$ measures the similarity between two uncertainty attributions. Following prior work on feature attribution evaluation~\cite{nauta2023anecdotal}, we quantify similarity between explanations using cosine similarity and Spearman's $\rho$, to measure both directional alignment and rank order correlation. Both metrics yield values in $[-1,1]$ with higher values indicating greater consistency.

\subsection{Continuity}
\label{prop:continuity}

Continuity refers to the smoothness of an explanation with respect to the inputs. The intuition behind continuity is that similar inputs should have similar explanations; in other \ac{xai} taxonomies, this aspect is also called robustness~\cite{ali_explainable_2023}. Iversen et al.~\cite{iversen2023identifying} evaluate the continuity of their uncertainty attribution method using the local Lipschitz continuity. For feature attributions, Agarwal et al.~\cite{agarwal_rethinking_2022} extend the local Lipschitz continuity by measuring the relative distance between explanations with respect to the distance between the inputs. We slightly adapt this metric for uncertainty attributions. Formally, the Relative Input Stability (RIS) is computed as follows:
\begin{align}
\mathrm{RIS}(x, x', u(x), u(x'))
= \max_{x'}\
&\frac{
\left\lVert \frac{u(x') - u(x)}{u(x)} \right\rVert_p
}{
\max\left(
\left\lVert \frac{x - x'}{x} \right\rVert_p,
\epsilon_{\min}\right)
}
, & \forall x' \text{ s.t. } x' \in \mathcal{N}_x,~~
{s}_x^2 \simeq {s}_{x'}^2 
\label{eq:ris}
\end{align}
The original definition of RIS in the context of feature attributions requires $\hat{y}_x = \hat{y}_{x'}$ for classification tasks. Since our uncertainty estimates are continuous, we do not require equality but rather that they are similar up to a small difference $\tau$, i.e.,
    $|s^2_{x}- s^2_{x'}| < \tau $.
We choose $\tau = 0.05$. In practice, we approximate \Cref{eq:ris} by generating $n_{pert}=50$ perturbations $x'$ of the input and keep only those for which ${s}_x^2 \simeq {s}_{x'}^2$. RIS values range between $[0,\infty)$. Lower RIS indicates smoother, more continuous attributions.

\subsection{Compactness}
\label{prop:compactness}

Motivated by human cognitive capacity limitations~\cite{longo2015designing}, compactness emphasises keeping explanations sparse, brief, and non-redundant, so they remain understandable~\cite{miller_explanation_2019,nauta2023anecdotal}. Our literature review revealed that the compactness property has not been evaluated in previous work on uncertainty attributions. In our evaluation framework, we use the complexity metric introduced by Bhatt et al.~\cite{bhatt_evaluating_2020} to evaluate the compactness of uncertainty attributions. The complexity of an uncertainty attribution $u$ is defined as the entropy of the attribution: 
\begin{align}
    \text{complexity}(u) = -\sum_{i=1}^n \mathbb{P}_u(i) \ln(\mathbb{P}_u(i)) \in [0,\ln(n)]
\end{align}
where $\mathbb{P}_u(i) = \dfrac{|u_i|}{\sum_{i=1}^n|u_i|}$ is the L1-normalised uncertainty attribution. Higher values indicate complexer attributions.

\subsection{Coherence and Conveyance}
\label{prop:conveyance}

Nauta et al.~\cite{nauta2023anecdotal} define coherence as the alignment of an explanation with domain knowledge or beliefs. This property is frequently evaluated in the context of uncertainty attribution. Patro et al.\cite{patro_u-cam_2019} compare uncertainty attributions to human-annotated uncertainty maps, while Brown and Talbert~\cite{brown_using_2022} calculate correlation between uncertainty and negative feature attributions, assuming uncertain features should negatively contribute to predictions. However, this assumption may not hold universally, as outlier features can simultaneously drive predictions and cause uncertainty. For example, consider a model trained to predict house prices, with size as a predictor on which the model relies heavily.
A house with a very large size, which could be considered an outlier within the training dataset, would yield an uncertain prediction — since that value is an outlier — yet the size would still be an important predictor.

Several papers evaluate uncertainty attributions by inducing distribution shifts in individual features and verifying whether the attributions correctly highlight the shifted features. Types of distribution shifts considered include covariate shifts~\cite{watson_explaining_2023}, \ac{ood} data or outliers~\cite{wang_semantic_2023,brown_using_2022,explainable-competency}, and controlled aleatoric noise~\cite{iversen2023identifying}. We collectively refer to these evaluation strategies as distribution shift detection. For example, Iversen et al.~\cite{iversen2023identifying} compute aleatoric uncertainty attributions for synthetic data with controlled noise and evaluate if their explanations highlight features with high ground-truth aleatoric uncertainty. Brown and Talbert~\cite{brown_using_2022} leverage a synthetic setup with known ground truth. They train a model on a synthetic tabular dataset with two strictly bounded features. They consider a test sample \ac{ood} if any of its features lie outside the training data boundaries. To evaluate their uncertainty attributions, they compute the fraction of \ac{ood} samples where the \ac{ood} feature had the highest uncertainty attribution. Note that this does not evaluate correctness, as it relies on prior knowledge and does not consider the model's internals. In summary, the evaluations with synthetic ground truth and distribution shift detection try to answer the same underlying question: 
\enquote{How reliably does uncertainty propagate to the uncertainty attributions based on prior beliefs and knowledge?}. 
We believe that this question is particularly important for uncertainty attributions and therefore propose \textbf{\textit{conveyance}} as an additional property specific to \ac{xuq}.

We define conveyance to measure how reliably uncertainty propagates to the uncertainty attributions. Closely related to coherence, it assesses whether attributions align with the expected sources of uncertainty, highlighting noisy or occluded regions, distribution-shifted features, or underrepresented training characteristics. To evaluate conveyance in our framework, we introduce two metrics: the Relative Rank Improvement (RRI), which is aligned with distribution-shift detection approaches in prior work, and a novel uncertainty-conveyance similarity (UCS) metric (Definition~\ref{def:ucs}).

\subsubsection{Relative Rank Improvement}
The core assumption behind RRI is that when a feature is perturbed to be out-of-distribution, become an outlier, or contain high aleatoric noise, its uncertainty attribution should increase correspondingly~\cite{brown_using_2022,wang_gradient-based_2023,explainable-competency,iversen2023identifying}. We define:
\begin{align}
    \text{RRI}(u(x_i), u(x'_i)) =  \frac{\text{rank}(u(x_i)) - \text{rank}(u(x'_i))}{\text{rank}(u(x_i))} 
    , & \enspace i=1,...,n
\end{align}

Here, $x'$ is a perturbed sample of an input $x$ such that feature $i$ is perturbed. Features are ranked by attribution magnitude in descending order, so $\text{rank}(u(x_i))=1$ indicates that feature $i$ contributes most to uncertainty. The \ac{rri} then compares the rank of the perturbed feature to its original rank, taking values in $[1-n,\frac{n-1}{n}]$. Positive RRI values indicate that the perturbed feature's rank has improved (i.e., moved toward higher importance), whereas negative values indicate degradation. For tabular data, we perturb feature $x_i$ to $x'_i = \mu_i+k \sigma_i$, where $\mu_i$ and $\sigma_i$ are the feature's mean and standard deviation in the training data. For each test sample, we perturb each feature individually and compute the average \ac{rri}. We set $k=4$. For images, we perturb pixel patches by inverting their colours (white pixels become black, and vice versa). We randomly perturb $25$ patches of each test image.

\subsubsection{Uncertainty Conveyance Similarity}
In addition to the \ac{rri}, we propose the \ac{ucs} metric to evaluate conveyance for ensemble-based uncertainty attribution methods, as proposed by Bley et al.~\cite{bley2025explaining} and previously described in~\Cref{section:preliminaries}. UCS compares empirical uncertainty attributions with an analytical approximation derived from first-order uncertainty propagation~\cite{chiaburu2025uncertainty}, extending beyond distribution-shift detection. In the following, we focus on \ac{mcd} as an \ac{uq} method and use predictive variance as a measure of uncertainty. 
To analytically approximate \ac{mcd}, we characterise a deterministic prediction model $f: (x,a) \mapsto \hat{y}$, where $\hat{y}\in \mathcal{Y} \subseteq \mathbb{R}$ in terms of the input $x \in \mathcal{X} \subseteq \mathbb{R}^n$ and its hidden layer activations $a \in \mathbb{R}^{l}$. $l$ denotes the number of neurons in the hidden layer. Let the explainer be a function $e: \mathbb{R}^{n +  l} \to \mathbb{R}^n$
mapping an input $x \in X \subseteq \mathbb{R}^n$ to a feature attribution explanation $e(x,a) \in \mathbb{R}^n$. 
To approximate explanation changes under stochastic forward passes with \ac{mcd}, we compute the Jacobian block $\mathbb{J}$ of the explainer with respect to the activations $( \mathbb{J})_{i,j} =  \frac{\partial e_i}{\partial a_j}; i = 1,..., ,n;  j = 1,...,l$.
Using the linearisation from Chiaburu et al.~\cite{chiaburu2025uncertainty}, we can model the feature attribution given a perturbation in the activations as: 
\begin{equation*}\label{eq:linearization}
    e(x,\hat{a}) \approx e(x,a) + \mathbb{J}\cdot \Delta a \eqqcolon e_{\text{lin}}(x)
\end{equation*}

For \ac{mcd}, we apply independent Bernoulli masks $B_j \sim \text{Ber}(p)$ to the activation vector $a$ (where $p$ is the dropout probability), yielding perturbations of the activations $\hat{a}_j = a_j (1-B_j)$. Thus, the change in activation after perturbation is $\Delta a_j = -a_j B_j$.  
Note that, unlike Chiaburu et al.~\cite{chiaburu2025uncertainty}, who add i.i.d. noise, our perturbations are activation-dependent: ${\Delta a}_j$ are independent across $j$ but not identically distributed, since their distributions scale with $a_j$.
The full perturbation vector $\Delta a$ can, therefore, be expressed as a linear transformation of the Bernoulli distributed random vector $B \in \mathbb{R}^l$ as follows:
\begin{equation*}
    \Delta a = AB =
    \begin{pmatrix}
        -a_1 & 0 & \dots & 0 \\
        0 & -a_2 & \dots & 0 \\
        \vdots & \vdots & \ddots & \vdots \\
        0 & 0 & \dots & -a_l
    \end{pmatrix}
    B
\end{equation*}

To simulate multiple \ac{mcd} forward passes and approximate the variance of $K$ feature attribution explanations, we can use the following equation:

\begin{equation}
    u_{\text{lin}}(x) \coloneqq \text{diag}(\mathrm{Var}(e_{\text{lin}}(x))) = \text{diag}(\mathbb{J} \cdot \Sigma_{\Delta a} \cdot \mathbb{J}^T)
    \label{eq:analytical_ua}
\end{equation}



The covariance of $\Delta a$ can be rewritten as $\Sigma_{\Delta a} = \Sigma_{AB} = A \cdot \Sigma_B \cdot A^T$. In summary, the uncertainty attributions of an input $x$ can be analytically approximated with \ac{mcd} as the underlying \ac{uq} method as follows\footnote{For \ac{mcdc}, the whole procedure works analogously by replacing the neuron activations with the weights.}:
\begin{equation*}\label{eq:var_lin}
    u_{lin}(x) = \text{diag}\left(\mathbb{J} \cdot p(1 - p) \cdot 
    \begin{pmatrix}
        a^2_1 & 0 & \dots & 0 \\
        0 & a^2_2 & \dots & 0 \\
        \vdots & \vdots & \ddots & \vdots \\
        0 & 0 & \dots & a^2_l
    \end{pmatrix}
    \cdot \mathbb{J}^T \right)\in \mathbb{R}^{n},
\end{equation*}
where $p(1-p)$ is the variance of a Bernoulli distributed variable.
Finally, to evaluate the original uncertainty attributions $u(x, \hat{y})$, we propose the following metric for conveyance:
\begin{definition}[Uncertainty conveyance similarity (UCS)]
Given an uncertainty attribution $u(x,\hat{y})$ and its analytical approximation $u_{\text{lin}}(x)$ as defined in~\Cref{eq:analytical_ua}, we define the uncertainty conveyance similarity as:
    \begin{align}
    \textnormal{UCS}(u(x,\hat{y})) = \textnormal{sim}(u(x,\hat{y}), u_{\text{lin}}(x)) 
    \end{align}
where $\textnormal{sim}(\cdot,\cdot)$ is a similarity metric. 
\label{def:ucs}
\end{definition}

Higher \ac{ucs} values signify better agreement with the analytical approximation and, relative to this baseline, better conveyance of uncertainty to the attributions. As for repeatability (see~\Cref{prop:consistency}), we use cosine similarity and Spearman's $\rho$, yielding values in $[-1,1]$.

    

%% file: experiments.tex
\section{Instantiation of the Evaluation Framework}
\label{section:experiments}

\begin{figure}[tb]
    \centering    \includegraphics[width=\columnwidth]{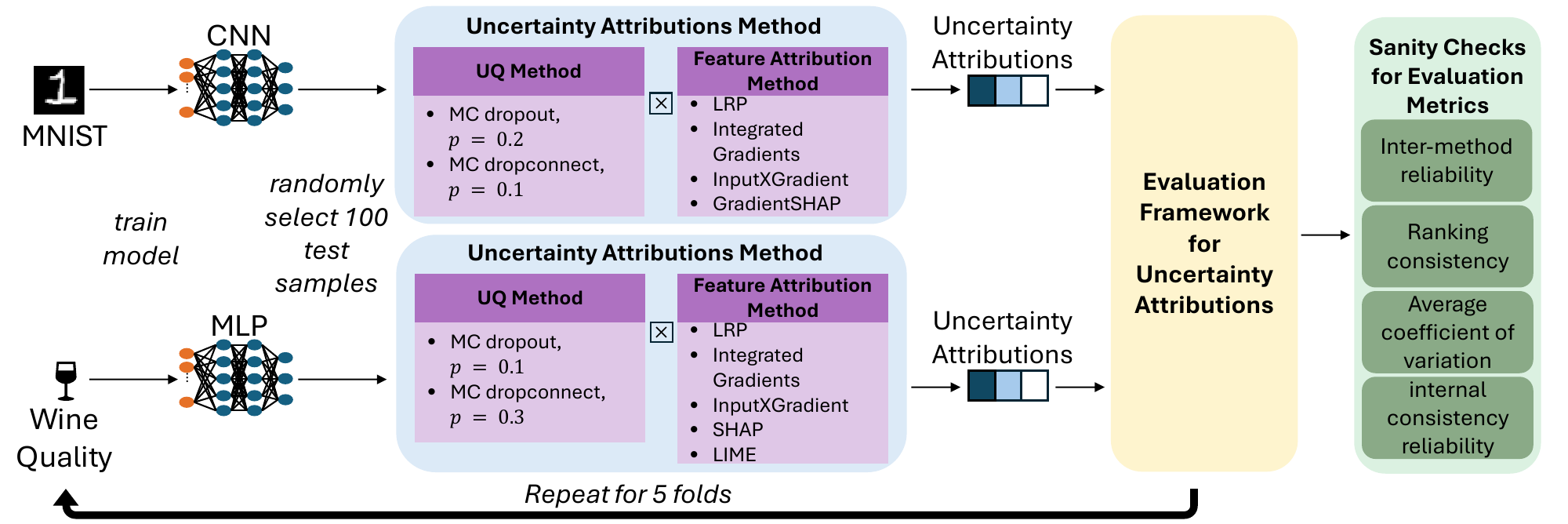}
    \caption{Experimental setup. Using 5-fold cross-validation, we train a CNN for MNIST and an MLP for Wine Quality. We compute uncertainty attributions for 100 test samples per fold using combinations of UQ and feature attribution methods (blue panel), rate them using our evaluation framework, and assess metrics via sanity checks.}
    \label{fig:experiment_setup}
\end{figure}

We demonstrate the evaluation framework by comparing multiple uncertainty attribution methods across two datasets. Each method combines an ensemble-based UQ technique (\ac{mcd} or \ac{mcdc}) with a feature attribution method. Figure~\ref{fig:experiment_setup} visualises the experimental setup. We apply the evaluation framework to two datasets. The first is the tabular Wine Quality dataset\footnote{Available at https://archive.ics.uci.edu/dataset/186/wine+quality}
, which comprises 11 continuous features. For this dataset, we use LRP~\cite{bach2015pixel}, Integrated Gradients (IG)~\cite{sundararajan2017axiomatic}, Input$\times$Gradient (IxG)~\cite{sundararajan2017axiomatic}, Shapely Value Sampling (SHAP)~\cite{lundberg_unified_2017}, and LIME~\cite{ribeiro_why_2016} for feature attribution. We deliberately include the stochastic methods SHAP and LIME to validate whether our proposed metrics adequately capture the theoretical limitations of sampling-based attribution approaches in the applied uncertainty attribution framework discussed in~\Cref{section:preliminaries}. Specifically, we hypothesize that if the evaluation metrics are effective, they should assign inferior scores to SHAP and LIME compared to the other~\ac{xai} methods. The second dataset is MNIST\footnote{Available at http://yann.lecun.com/exdb/mnist}. For MNIST, we use four methods: LRP, IG, IxG, and GradientSHAP~\cite{lundberg_unified_2017}. We use implementations from the zennit~\cite{anders_software_2026} and captum\footnote{https://captum.ai/} 
libraries.

We assess metric reliability via four sanity checks:
\begin{enumerate*}[label=(\roman*)]
    \item inter-method reliability~\cite{tomsett_sanity_2020},
    \item ranking consistency,
    \item average coefficient of variation, and
    \item internal consistency reliability.
\end{enumerate*}

Inter-method reliability measures the agreement between different uncertainty attribution methods via pairwise Spearman's $\rho$ of their scores across samples~\cite{tomsett_sanity_2020}. Ranking consistency measures how stable methods rank relative to each other via pairwise Spearman's $\rho$ of samples' scores across methods. Both metrics range in $[-1,1]$, with higher values indicating better agreement and consistency. The average coefficient of variation quantifies score dispersion across samples relative to the mean, averaged across all methods for each metric. Additionally, we compute the internal consistency reliability specifically for the two conveyance metrics (RRI and UCS) to assess whether they measure the same underlying construct, in line with~\cite{tomsett_sanity_2020}. Internal consistency reliability is computed as Spearman's $\rho$ between scores from two metrics for attributions from the same uncertainty attribution method. This check applies only to conveyance, as it is the only property operationalised across more than one metric, and assessing the correlation between the newly introduced UCS and RRI indicates whether UCS provides analogous or complementary information.

For Wine Quality, we train a feedforward network with two hidden layers of 50 neurons for regression. For MNIST classification, we train a CNN with two convolutional blocks followed by two linear layers with $50$ neurons each. We use 5-fold cross-validation for both tasks to mitigate sampling bias and provide robust metric estimates. \ac{mcd} is enabled for all linear layers; due to computational constraints, \ac{mcdc} is applied only to the final layer. We tune drop probability $p \in [0.1,0.2,...,0.5]$ via grid search, jointly optimising MSE, NLL, and CRPS (regression) or Brier score (classification). For Wine Quality, we selected $p=0.1$ for \ac{mcd} as the best trade-off across metrics, and $p=0.3$ for \ac{mcdc}. For MNIST, we chose $p=0.2$ for \ac{mcd} and $p=0.1$ for dropconnect. To evaluate the uncertainty attribution methods, we report each metric from~\Cref{section:properties} for $100$ randomly selected samples per fold.

%% file: results_new.tex
\section{Results}
\label{section:results}

In this section, we present the experimental results. We compare the performance of the various combinations of \ac{uq} and feature attribution methods. Further, we analyse the reliability of the proposed metrics via established sanity checks.

\subsection{Evaluation of Uncertainty Attribution Methods}
\label{results}

\begin{figure}[h]
    \centering    \includegraphics[width=\columnwidth]{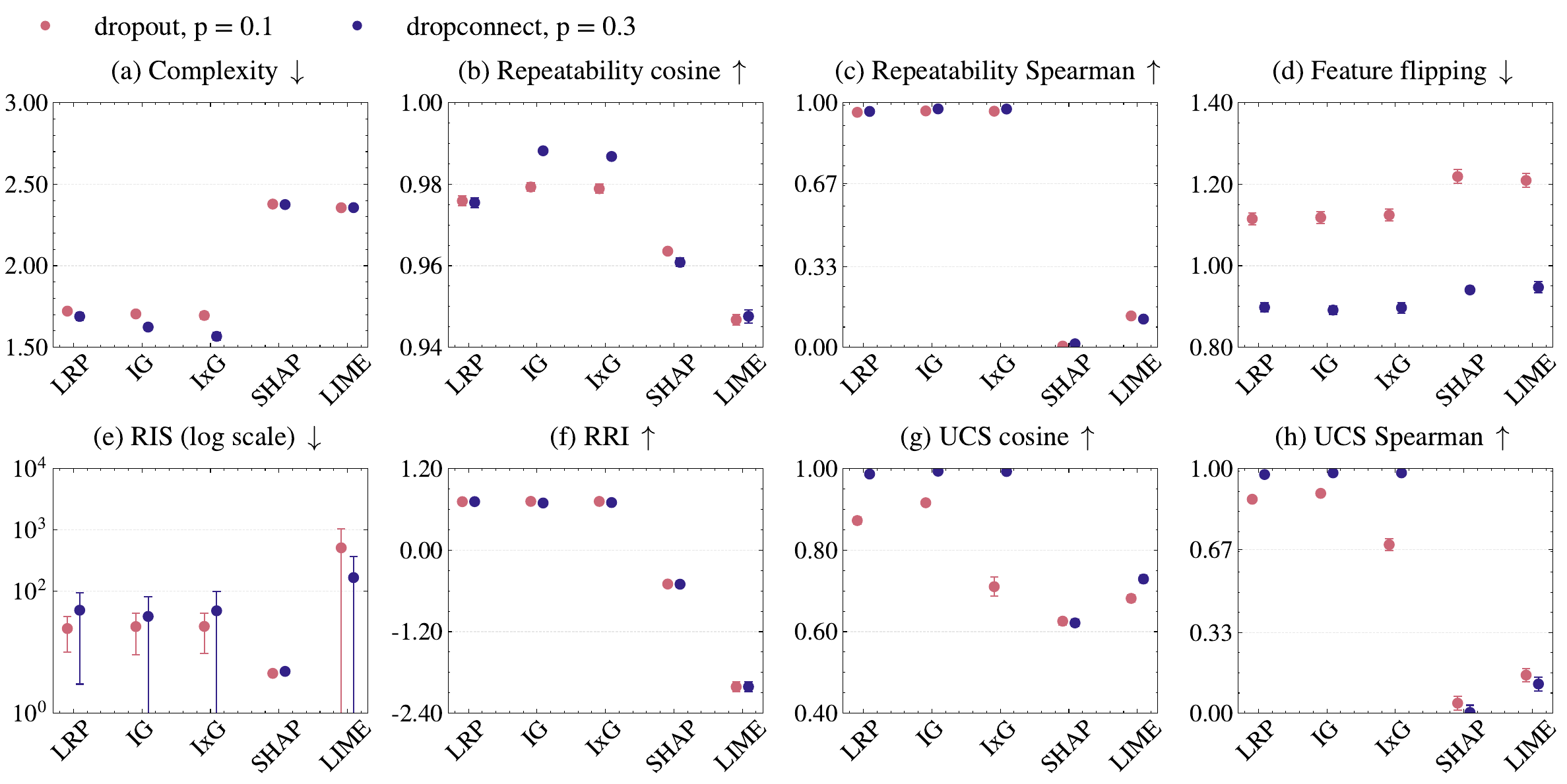}
    \caption{Metric scores for the Wine Quality dataset with 95\% confidence interval. Arrows indicate whether lower ($\downarrow$) or higher ($\uparrow$) values are better. Gradient-based methods (IxG, IG) and LRP outperform perturbation-based methods on most metrics for \ac{mcd} and \ac{mcdc}. \ac{mcdc} shows better repeatability, feature flipping, and higher UCS than MCD.}
    \label{fig:wine_quality_dotplots}
\end{figure}

\begin{figure}[h]
    \centering    \includegraphics[width=\columnwidth]{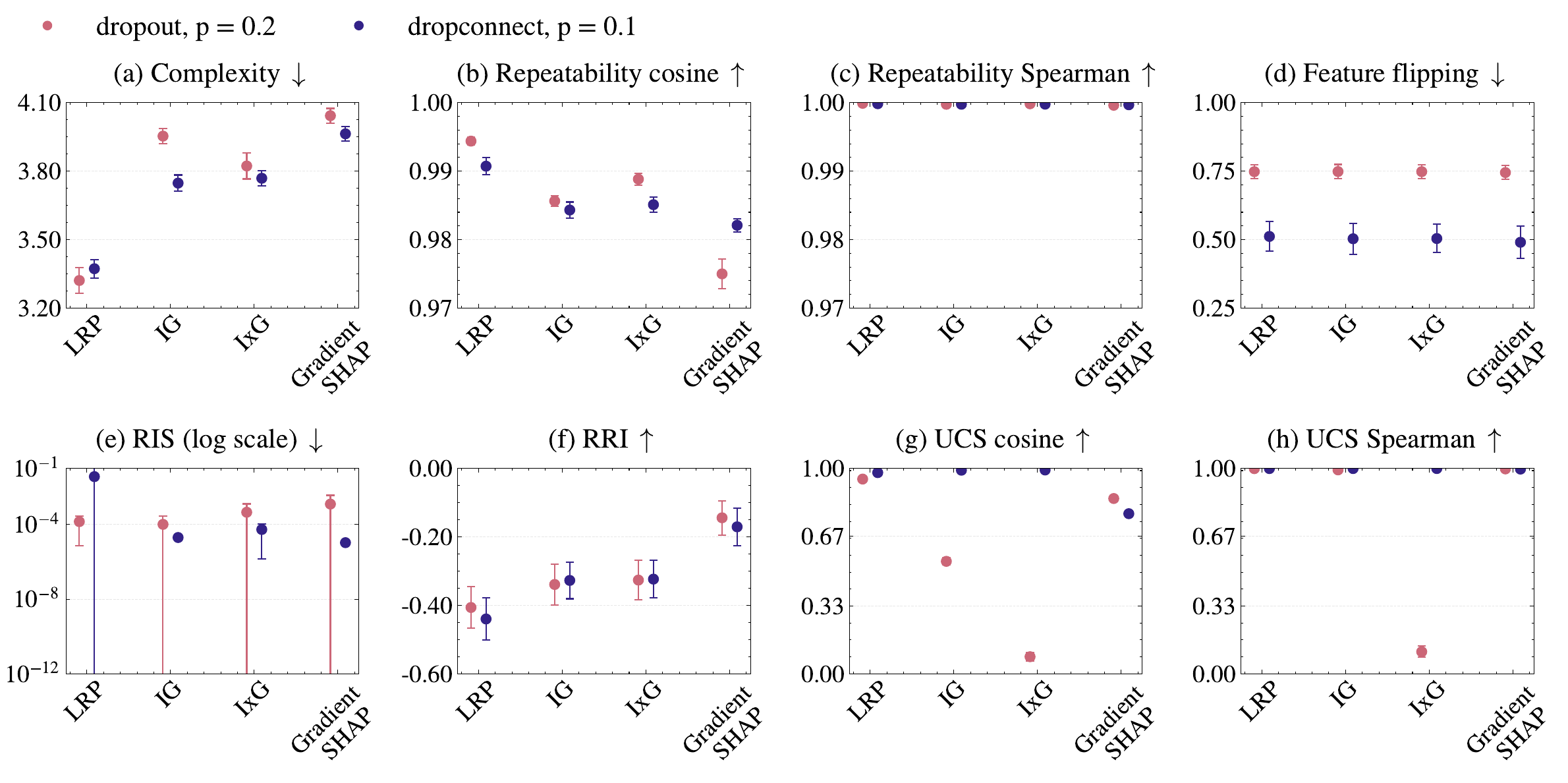}
    \caption{Metric scores for the MNIST dataset with 95\% confidence interval. Arrows indicate whether lower ($\downarrow$) or higher ($\uparrow$) values are better. \ac{mcdc} outperforms \ac{mcd} in UCS but achieves higher feature flipping scores. All methods exhibit near-perfect repeatability and RIS.}
    \label{fig:mnist_dotplots}
\end{figure}

Examples of uncertainty and feature attributions are shown in~\Cref{fig:example_attributions}.~\Cref{tab:metric_scores_methods_combined} summarises the metric scores for all uncertainty attribution methods on the Wine Quality and MNIST datasets, averaged over all 5 folds. Additionally,~\Cref{fig:wine_quality_dotplots} and~\Cref{fig:mnist_dotplots} visualise the metric scores on both datasets. 

\begin{table}[t]
\renewcommand{\arraystretch}{1.2}
\caption{Metric scores averaged over all 5 folds for Wine Quality and MNIST datasets. Arrows indicate whether lower ($\downarrow$) or higher ($\uparrow$) values are better. Best metric values across all uncertainty attribution methods are highlighted in blue.}
\setlength{\tabcolsep}{4pt}
\begin{tabular}{c p{1.8cm} p{1.8cm} p{1.2cm} p{1.2cm} p{1.2cm} p{1.2cm} p{1.2cm}}
\hline
& \textbf{Metric} & \textbf{UQ Method} & \textbf{LRP} & \textbf{IG} & \textbf{IxG} & \textbf{SHAP} & \textbf{LIME} \\
\hline
\multirow{16}{*}{\begin{sideways}\textbf{Wine Quality}\end{sideways}}
& \multirow{2}{*}{complexity $\downarrow$} & \ac{mcd} & 1.722 & 1.704 & 1.695 & 2.378 & 2.356 \\
& & \ac{mcdc} & 1.689 & 1.624 & \cellcolor{blue!25}\textbf{1.567} & 2.375 & 2.357 \\
\hhline{~-------}
& \multirow{2}{*}{\shortstack[l]{repeatability \\ (cosine) $\uparrow$}} & \ac{mcd} & 0.976 & 0.979 & 0.979 & 0.964 & 0.947 \\
& & \ac{mcdc} & 0.976 & \cellcolor{blue!25}\textbf{0.988} & 0.987 & 0.961 & 0.948 \\
\hhline{~-------}
& \multirow{2}{*}{\shortstack[l]{repeatability \\ (Spearman) $\uparrow$}} & \ac{mcd} & 0.961 & 0.967 & 0.966 & 0.005 & 0.128 \\
& & \ac{mcdc} & 0.965 & \cellcolor{blue!25}\textbf{0.975} & 0.974 & 0.014 & 0.115 \\
\hhline{~-------}
& \multirow{2}{*}{\shortstack[l]{feature \\ flipping $\downarrow$}} & \ac{mcd} & 1.115 & 1.119 & 1.124 & 1.219 & 1.209 \\
& & \ac{mcdc} & 0.898 & \cellcolor{blue!25}\textbf{0.891} & 0.897 & 0.941 & 0.947 \\
\hhline{~-------}
& \multirow{2}{*}{RIS $\downarrow$} & \ac{mcd} & 24.1 & 25.9 & 26.2 & \cellcolor{blue!25}\textbf{4.5} & 510.4 \\
& & \ac{mcdc} & 45.9 & 35.3 & 43.5 & 4.9 & 172.8 \\
\hhline{~-------}
& \multirow{2}{*}{RRI $\uparrow$} & \ac{mcd} & 0.715 & \cellcolor{blue!25}\textbf{0.719} & \cellcolor{blue!25}\textbf{0.719} & -0.498 & -2.012 \\
& & \ac{mcdc} & 0.715 & 0.695 & 0.702 & -0.501 & -2.011 \\
\hhline{~-------}
& \multirow{2}{*}{\shortstack[l]{UCS \\ (cosine) $\uparrow$}} & \ac{mcd} & 0.873 & 0.916 & 0.711 & 0.626 & 0.682 \\
& & \ac{mcdc} & 0.987 & \cellcolor{blue!25}\textbf{0.994} & 0.993 & 0.622 & 0.729 \\
\hhline{~-------}
& \multirow{2}{*}{\shortstack[l]{UCS \\ (Spearman) $\uparrow$}} & \ac{mcd} & 0.875 & 0.899 & 0.689 & 0.041 & 0.156 \\
& & \ac{mcdc} & 0.976 & \cellcolor{blue!25}\textbf{0.983} & \cellcolor{blue!25}\textbf{0.983} & 0.003 & 0.119 \\
\hline
& \textbf{Metric} & \textbf{UQ Method} & \textbf{LRP} & \textbf{IG} & \textbf{IxG} & \multicolumn{2}{c}{\textbf{GradientSHAP}}  \\
\hline
\multirow{16}{*}{\begin{sideways}\textbf{MNIST}\end{sideways}}
& \multirow{2}{*}{complexity $\downarrow$} & \ac{mcd} & \cellcolor{blue!25}\textbf{3.321} & 3.954 & 3.823 & \multicolumn{2}{c}{4.043} \\
& & \ac{mcdc} & 3.372 & 3.748 & 3.769 & \multicolumn{2}{c}{3.964} \\
\hhline{~-------}
& \multirow{2}{*}{\shortstack[l]{repeatability \\ (cosine) $\uparrow$}} & \ac{mcd} & \cellcolor{blue!25}\textbf{0.994} & 0.986 & 0.989 & \multicolumn{2}{c}{0.975} \\
& & \ac{mcdc} & 0.991 & 0.984 & 0.985 & \multicolumn{2}{c}{0.982} \\
\hhline{~-------}
& \multirow{2}{*}{\shortstack[l]{repeatability \\ (Spearman) $\uparrow$}} & \ac{mcd} & \cellcolor{blue!25}\textbf{1.000} & \cellcolor{blue!25}\textbf{1.000} & \cellcolor{blue!25}\textbf{1.000} & \multicolumn{2}{c}{\cellcolor{blue!25}\textbf{1.000}} \\
& & \ac{mcdc} & \cellcolor{blue!25}\textbf{1.000} & \cellcolor{blue!25}\textbf{1.000} & \cellcolor{blue!25}\textbf{1.000} & \multicolumn{2}{c}{\cellcolor{blue!25}\textbf{1.000}} \\
\hhline{~-------}
& \multirow{2}{*}{\shortstack[l]{feature \\ flipping $\downarrow$}} & \ac{mcd} & 0.748 & 0.749 & 0.748 & \multicolumn{2}{c}{0.745} \\
& & \ac{mcdc} & 0.512 & 0.503 & 0.504 & \multicolumn{2}{c}{\cellcolor{blue!25}\textbf{0.490}} \\
\hhline{~-------}
& \multirow{2}{*}{RIS $\downarrow$} & \ac{mcd} & \cellcolor{blue!25}\textbf{0.000} & \cellcolor{blue!25}\textbf{0.000} & \cellcolor{blue!25}\textbf{0.000} & \multicolumn{2}{c}{0.001} \\
& & \ac{mcdc} & 0.037 & \cellcolor{blue!25}\textbf{0.000} & \cellcolor{blue!25}\textbf{0.000} & \multicolumn{2}{c}{\cellcolor{blue!25}\textbf{0.000}} \\
\hhline{~-------}
& \multirow{2}{*}{RRI $\uparrow$} & \ac{mcd} & -0.406 & -0.339 & -0.326 & \multicolumn{2}{c}{\cellcolor{blue!25}\textbf{-0.144}} \\
& & \ac{mcdc} & -0.440 & -0.327 & -0.323 & \multicolumn{2}{c}{-0.170} \\
\hhline{~-------}
& \multirow{2}{*}{\shortstack[l]{UCS \\ (cosine) $\uparrow$}} & \ac{mcd} & 0.949 & 0.548 & 0.083 & \multicolumn{2}{c}{0.854} \\
& & \ac{mcdc} & 0.980 & 0.992 & \cellcolor{blue!25}\textbf{0.993} & \multicolumn{2}{c}{0.780} \\
\hhline{~-------}
& \multirow{2}{*}{\shortstack[l]{UCS \\ (Spearman) $\uparrow$}} & \ac{mcd} & 0.999 & 0.993 & 0.108 & \multicolumn{2}{c}{0.998} \\
& & \ac{mcdc} & \cellcolor{blue!25}\textbf{1.000} & \cellcolor{blue!25}\textbf{1.000} & \cellcolor{blue!25}\textbf{1.000} & \multicolumn{2}{c}{0.997} \\
\hline
\end{tabular}
\label{tab:metric_scores_methods_combined}
\end{table}

\paragraph{Gradient-based Methods and LRP Outperform Perturbation-based Approaches on Wine Quality. } Perturbation-based uncertainty attribution methods (LIME, SHAP) combined with both \ac{mcd} and \ac{mcdc} achieve notably lower metric scores than gradient-based methods (IG, IxG) and LRP on the Wine Quality dataset (\Cref{fig:wine_quality_dotplots} and~\Cref{tab:metric_scores_methods_combined}). For instance, LRP, IG, and IxG demonstrate consistently high repeatability, as opposed to LIME and SHAP (\Cref{fig:wine_quality_dotplots} (b)-(c)). A similar trend is observed for the UCS score: both SHAP and LIME show notably lower values than LRP and IG, with IxG also showing lower values for MCD (\Cref{fig:wine_quality_dotplots} (g)-(h)). The RRI metric further supports this finding. IG, LRP, and IxG achieve the highest scores, while LIME and SHAP exhibit negative values, 
indicating that perturbed features expected to have high uncertainty attribution actually moved down in rank. Regarding complexity, IxG achieves the lowest scores (\Cref{fig:wine_quality_dotplots} (a) indicating sparser attributions, while SHAP and LIME produce more complex attributions. For correctness, all \ac{mcdc} and all \ac{mcd} methods exhibit similar performance. IG demonstrates the best feature flipping performance (for \ac{mcdc}). Interestingly, despite its stochasticity, SHAP with both \ac{uq} methods achieves the lowest RIS scores (\Cref{fig:wine_quality_dotplots} (e)).

\paragraph{High Consistency but Varying Conveyance on MNIST. } On MNIST, the four attribution methods exhibit less variability in metric scores compared to Wine Quality (\Cref{fig:mnist_dotplots} and~\Cref{tab:metric_scores_methods_combined}). We excluded methods based purely on perturbation here due to computational constraints. Repeatability on MNIST is very high, with all methods achieving near-perfect Spearman's $\rho$ ($\sim1.0$) for both \ac{mcd} and \ac{mcdc}, and cosine similarities above 0.975 (\Cref{fig:mnist_dotplots} (b)-(c)). Unlike Wine Quality, the RIS metric achieves near-zero values across all method combinations (\Cref{fig:mnist_dotplots} (e)). However, the RRI exhibits negative values for all methods, indicating that the modified high-uncertainty features unexpectedly decreased in uncertainty attribution (\Cref{fig:mnist_dotplots} (f)). UCS (cosine) values are substantially lower than on Wine Quality for \ac{mcd} with IG and IxG; however, all \ac{mcdc} combinations except IxG achieve UCS Spearman scores close to 1.0 (\Cref{fig:mnist_dotplots} (g)-(h)). 

\paragraph{The Choice of UQ Method Notably Impacts Metric Scores. } On both datasets, \ac{mcdc} improves correctness across all methods (\Cref{fig:wine_quality_dotplots} (d)). The UCS metric shows substantial improvements across most methods with \ac{mcdc}, as the near-perfect scores for LRP, IG, and IxG in~\Cref{fig:wine_quality_dotplots} (g)-(h) and \Cref{fig:mnist_dotplots} (g)-(h) reveal.
On the Wine Quality dataset, the RIS metric for LRP, IG, and IxG shows a small degradation when using \ac{mcdc} instead of \ac{mcd}. With \ac{mcdc}, RIS also exhibits a particularly broad confidence interval~\Cref{fig:wine_quality_dotplots} (e). 

\begin{table}[t]
\renewcommand{\arraystretch}{1.3} 

\setlength{\tabcolsep}{8pt} 
\caption{Sanity checks of all evaluation metrics for Wine Quality and MNIST. Arrows indicate whether lower ($\downarrow$) or higher ($\uparrow$) values are better for each sanity check. Best scores for Wine Quality and MNIST are highlighted in blue and red, respectively.}
\begin{tabular}{lllll}
\hline
\textbf{Metric}                         & \textbf{Dataset}      & \begin{tabular}[c]{@{}l@{}}\textbf{Inter-method} \\ \textbf{reliability} $\uparrow$\end{tabular} & \begin{tabular}[c]{@{}l@{}}\textbf{Ranking} \\ \textbf{consistency} $\uparrow$\end{tabular} & \begin{tabular}[c]{@{}l@{}}\textbf{Average coefficient} \\ \textbf{of variation}  $\downarrow$\end{tabular} \\
\hline
\multirow{2}{*}{complexity}    & Wine Quality & 0.279                    & 0.766             &\cellcolor{blue!25}\textbf{0.018}      \\
                               & MNIST        &\cellcolor{red!25}\textbf{0.656}                     & 0.553             & 0.135     \\
\hline
\multirow{2}{*}{\begin{tabular}[c]{@{}l@{}}repeatability\\ (cosine)\end{tabular}} & Wine Quality & 0.076              & 0.646             & 0.019    \\
                               & MNIST        & 0.213                    & 0.328             & 0.014     \\
\hline
\multirow{2}{*}{\begin{tabular}[c]{@{}l@{}}repeatability\\ (Spearman)\end{tabular}} & Wine Quality & 0.168            & 0.762             & 1.227     \\
                               & MNIST        & 0.647                   & 0.527             & \cellcolor{red!25}\textbf{0.000}    \\
\hline
\multirow{2}{*}{\begin{tabular}[c]{@{}l@{}}feature\\ flipping\end{tabular}} & Wine Quality & \cellcolor{blue!25}\textbf{0.821}                & 0.845             & 0.166     \\
                               & MNIST        & 0.036                    & 0.261             & 1.200    \\
\hline
\multirow{2}{*}{RIS}           & Wine Quality & 0.074                    & 0.136             & 8.858     \\
                               & MNIST        & 0.399                    & 0.280             & 21.577     \\
\hline
\multirow{2}{*}{RRI}           & Wine Quality & 0.293                    & 0.618             & 1.404     \\
                               & MNIST        & 0.503                   & 0.025             & 8.064    \\
\hline
\multirow{2}{*}{\begin{tabular}[c]{@{}l@{}}UCS\\ (cosine)\end{tabular}}   & Wine Quality & 0.021                    & 0.787             & 0.146     \\
                               & MNIST        & 0.048                    & 0.787             & 0.054     \\
\hline
\multirow{2}{*}{\begin{tabular}[c]{@{}l@{}}UCS\\ (Spearman)\end{tabular}} & Wine Quality & 0.064                    & \cellcolor{blue!25}\textbf{0.862}             & 2.646     \\
                               & MNIST        & 0.410                    & \cellcolor{red!25}\textbf{0.847}            & \cellcolor{red!25}\textbf{0.000}     \\
\hline
\end{tabular}
\label{tab:sanity_checks}
\end{table}

\begin{table}[b]
\renewcommand{\arraystretch}{1.3} 
\caption{Spearman's $\rho$ (internal consistency reliability) between pairs of conveyance metrics for each uncertainty attribution method for Wine Quality and MNIST.}
\setlength{\tabcolsep}{4pt} 
\begin{tabular}{p{0.4cm} l l l l l l l}
\hline
& \textbf{Metrics} & \textbf{UQ Method} & \textbf{LRP} & \textbf{IG} & \textbf{IxG} & \textbf{SHAP} & \textbf{LIME} \\
\hline
\multirow{4}{*}{\begin{sideways}\multirow{2}{*}{\shortstack[c]{\textbf{Wine} \\ \textbf{Quality}}}\end{sideways}}
& \multirow{2}{*}{\shortstack[l]{UCS (Spearman) \\ vs RRI}} & \ac{mcd} & 0.025 & 0.045 & 0.087 & -0.015 & -0.002 \\
& & \ac{mcdc} & 0.032 & -0.003 & 0.034 & 0.049 & -0.001 \\
\cline{2-8}
& \multirow{2}{*}{\shortstack[l]{UCS (cosine) \\ vs RRI}} & \ac{mcd} & -0.100 & 0.038 & -0.106 & 0.014 & -0.002 \\
& & \ac{mcdc} & -0.031 & 0.031 & 0.011 & 0.027 & -0.004 \\
\hline
& \textbf{Metrics} & \textbf{UQ Method} & \textbf{LRP} & \textbf{IG} & \textbf{IxG} & \multicolumn{2}{c}{\textbf{GradientSHAP}}  \\
\hline
\multirow{4}{*}{\begin{sideways}\multirow{2}{*}{\shortstack[c]{\textbf{MNIST}}}\end{sideways}}
& \multirow{2}{*}{\shortstack[l]{UCS (Spearman) \\ vs RRI}} & \ac{mcd} & 0.025 & -0.011 & -0.024 & \multicolumn{2}{c}{0.001} \\
& & \ac{mcdc} & 0.000 & 0.023 & 0.016 & \multicolumn{2}{c}{-0.004} \\
\cline{2-8}
& \multirow{2}{*}{\shortstack[l]{UCS (cosine) \\ vs RRI}} & \ac{mcd} & 0.041 & 0.094 & -0.021 &\multicolumn{2}{c}{0.007} \\
& & \ac{mcdc} & -0.043 & 0.004 & 0.022 & \multicolumn{2}{c}{0.000} \\
\hline

\end{tabular}
\label{tab:internal_consistency_reliability}
\end{table}

\subsection{Sanity Checks for Evaluation Metrics}
\Cref{tab:sanity_checks} shows the results of the inter-method reliability, ranking consistency, and average coefficient of variation per metrics and dataset. Feature flipping achieves the highest inter-method reliability for Wine Quality (0.821), while complexity and repeatability (Spearman) show high agreement on MNIST (0.656 and 0.647). Most other metrics show low inter-method reliability, with values below 0.5, indicating that different uncertainty attribution methods yield substantially different sample rankings. This is particularly evident for feature flipping on MNIST and UCS (cosine). Several metrics demonstrate high ranking consistency on Wine Quality: complexity (0.766), repeatability (Spearman) (0.762), UCS (cosine) (0.787), and UCS (Spearman) (0.862). This indicates that these metrics produce stable rankings of uncertainty attribution methods largely independent of the sample. On MNIST, UCS metrics exhibit similar results (0.787 and 0.847), whereas other metrics exhibit greater variability across samples. RIS shows low ranking consistency on Wine Quality (0.136), and RRI shows near-zero ranking consistency on MNIST, suggesting high sample-dependence. Most metrics exhibit relatively low variation on Wine Quality, except UCS (Spearman) (2.646) and RIS (8.858). On MNIST, the pattern is more extreme: while most metrics show very low variation, RIS exhibits exceptionally high dispersion (21.577). RIS exhibits high variability across both datasets. This high variability may indicate sensitivity to sample-specific characteristics or instability in the metric. As shown in~\Cref{tab:internal_consistency_reliability}, all uncertainty attribution methods exhibit low internal consistency reliability, with near-zero Spearman's $\rho$ on both datasets and across both UCS metrics, compared to RRI.

%% file: discussion_new.tex
\section{Discussion}
\label{section:discussion}

\paragraph{Gradient-based Methods and LRP Align with Theoretical Requirements. }
The superiority of gradient-based methods and LRP over perturbation-based approaches on the Wine Quality dataset aligns with the theoretical constraints of the uncertainty attribution approach. As hypothesized in the experimental design, the metrics successfully identify Shapley Value Sampling (SHAP) and LIME as inferior to the deterministic methods. However, the feature flipping scores are only marginally higher than those of the conforming attribution methods, IG, IxG, and LRP (\Cref{fig:wine_quality_dotplots} (d)). A notable exception is RIS, which fails to capture the theoretical limitations of stochastic sampling, as SHAP achieves the lowest RIS score on Wine Quality (\Cref{fig:wine_quality_dotplots} (e)). SHAP and LIME's poor UCS scores confirm the theoretical mismatch, as UCS explicitly measures consistency with linear approximations. 

\paragraph{Dataset Characteristics Strongly Influence Metric Behaviour. }

MNIST and Wine Quality reveal notably different patterns, highlighting the role of data and task characteristics. MNIST exhibits near-zero RIS scores, whereas RIS scores on Wine Quality are rather high. On Wine Quality, on the other hand, feature flipping scores are rather high. We attribute these differences to the relatively simple patterns in the data, which result in low ensemble variance. When uncertainty estimates and attributions are small in absolute value, minor perturbation-induced changes may produce large relative changes, thereby artificially inflating feature flipping scores. This reveals a fundamental limitation of deletion-style correctness metrics. Setting features to baseline values may shift inputs outside the training distribution, causing unpredictable (epistemic) uncertainty. Removing high-uncertainty features may introduce new sources of uncertainty rather than reducing it. Feature flipping's coefficient of variation above 1 and its low inter-method reliability and ranking consistency on MNIST (\Cref{tab:sanity_checks}) further cast doubt on its trustworthiness. In summary, this reflects established concerns regarding faithfulness metrics for feature attributions~\cite{tomsett_sanity_2020}, for which different approaches have been proposed~\cite{zheng2024f}. MNIST's perfect repeatability and near-zero RIS suggest either insufficient metric sensitivity for high-dimensional data, or genuinely deterministic attribution behaviour despite stochastic \ac{uq} methods. Future work should explore alternative correctness metrics that address these limitations, disentangle metric limitations from inherent attribution properties across modalities, and recognise that results may not generalise across datasets with different characteristics.

\paragraph{Metric-Specific Limitations Require Careful Interpretation. }

Poor UCS (cosine) scores for \ac{mcd} on both datasets compared to \ac{mcdc} show metric validity boundaries (\Cref{tab:metric_scores_methods_combined}). The analytical approximation underlying UCS depends on activation (\ac{mcd}) or weight (\ac{mcdc}) perturbations (see~\Cref{def:ucs}). For \ac{mcd}, activation values can attain large absolute magnitudes, potentially rendering the first-order linear approximation unreliable. Additionally, computational constraints limited \ac{mcdc} to the final layer, whereas \ac{mcd} was applied to all layers, which may explain the observed performance differences. A limitation of Spearman's $\rho$ that we experienced during the experiments is that it is undefined for uniform vectors. We set Spearman's $\rho$ to 0 for non-identical uniform vectors and 1 for identical vectors. On MNIST, analytical approximations frequently yielded all-zero attributions, potentially artificially lowering UCS scores. RIS exhibits a high variation in metric scores on Wine Quality (\Cref{fig:wine_quality_dotplots} (e)). Individual-sample analysis showed that a few extreme outliers drive the high mean, as evidenced by the high coefficient of variation (\Cref{tab:sanity_checks}). This suggests that the overall average is an inappropriate statistic for RIS. RRI exhibits particularly poor performance on MNIST (negative values values -~\Cref{fig:mnist_dotplots} (f) and~\Cref{tab:metric_scores_methods_combined}). This raises questions about whether it reflects poor attribution quality or limitations in metric design for high-dimensional data, potentially due to a suboptimal perturbation strategy. One avenue for alleviating this problem is to use metrics based on manifold-based perturbations, such as the one introduced by Vascotto et al.~\cite{vascotto2025can} for feature attributions.

\paragraph{Practical Implications for Evaluation and Method Selection. }

The sanity checks (\Cref{tab:sanity_checks}) provide guidance for metric selection in future evaluations. Metrics with high ranking consistency, complexity, and repeatability (Spearman), and both UCS metrics, are particularly suitable for comparing and selecting among uncertainty attribution methods, as they yield stable rankings that are independent of specific samples. This consistency indicates that relative performance generalises across inputs within a dataset. In contrast, low inter-method reliability does not necessarily indicate metric failure. Most metrics exhibit inter-method reliability below 0.5, indicating that different uncertainty attribution methods produce substantially different sample-level rankings. This aligns with \ac{xai} research on faithfulness metrics~\cite{tomsett_sanity_2020}, in which low agreement is expected when methods employ different mechanisms and capture different behavioural aspects. However, metrics combining low ranking consistency with high coefficient of variation, particularly RRI, RIS, and feature flipping on MNIST (\Cref{tab:sanity_checks}) should be interpreted with caution. They are highly sample-dependent and may not provide reliable indications of overall method quality. 

Furthermore, the low absolute internal consistency reliability across uncertainty attributions between RRI and both UCS metrics (see~\Cref{tab:internal_consistency_reliability}) indicates that they capture distinct facets of conveyance. By design, RRI evaluates alignment with domain-based prior beliefs about which features should contribute to uncertainty (e.g., out-of-distribution or outlier features), while UCS assesses technical propagation consistency with a first-order approximation of the \ac{uq} mechanism. The lack of correlation between these metrics is thus not problematic, but rather indicates that they provide complementary information about different aspects of the same property. Overall, the sanity-check results (\Cref{tab:sanity_checks} and~\Cref{tab:internal_consistency_reliability}) support a multi-metric evaluation approach, as no single metric captures all relevant aspects of uncertainty attribution quality. Note that the concrete selection of properties and metrics depends on the specific use case. For instance, compactness might be rather irrelevant for image data. However, its importance increases when dealing with high-dimensional tabular data.

\paragraph{Limitations. } 
The evaluation framework is applied to two datasets, with a sample size of 100 instances per fold, constrained by computational resources. This may affect the generalisability of the findings regarding the metrics' reliability. The experiments serve as illustrations of how the framework can be applied, rather than as generalisable conclusions. Replication across additional datasets, domains, and data modalities is necessary to establish the broader applicability of the findings. Additionally, the experiments focus on the uncertainty attribution approach from Bley et al.~\cite{bley2025explaining}. By design, the proposed UCS metric is specifically tailored to this framework and may not directly transfer to alternative approaches. Furthermore, hyperparameter sensitivity (baseline selection, perturbation strategies, similarity metrics) of metrics and methods represents an important direction for future work. While we followed standard practices from \ac{xai} and \ac{xuq} literature~\cite{agarwal_rethinking_2022,bley2025explaining,perez_attribution_2022,nauta2023anecdotal}, a systematic analysis could improve metric reliability. The experimental results reveal additional limitations. Although experimental results show that most metrics appropriately penalize SHAP and LIME, feature flipping exhibits only marginally higher scores than IG, IxG, and LRP. Combined with feature flipping's poor sanity check results on MNIST, this underscores the need for novel correctness metrics specifically designed for uncertainty attributions. Finally, our work currently operationalises only a subset of the Co-12 properties proposed in~\cite{nauta2023anecdotal} and addressing the remaining ones could enhance the evaluation of uncertainty attribution methods.

%% file: conclusion.tex
\section{Conclusion}
\label{section:conclusion}
Understanding uncertainty in AI predictions is essential for deploying reliable systems in critical applications, yet explaining \textit{why} a model is uncertain remains underexplored. Uncertainty attribution methods address this gap by identifying which input features influence prediction uncertainty. However, their evaluation varies across studies. Studies typically employ heterogeneous proxy tasks and metrics for evaluation, which limit meaningful comparisons of methods and the assessment of quality. 
This work establishes a structured evaluation framework for uncertainty attribution methods by adapting four Co-12 properties~\cite{nauta2023anecdotal} (correctness, consistency, continuity, and compactness). Additionally, we introduced \emph{conveyance}, a novel \ac{xuq}-specific property, and developed a corresponding metric, uncertainty conveyance similarity (UCS). Through empirical experiments, we demonstrated that the framework provides guidance for analysing and selecting appropriate uncertainty attribution methods. We further showed that UCS and RRI, both assessing conveyance, exhibit low absolute internal consistency reliability. This indicates that they measure different aspects of conveyance. The introduction of UCS, therefore, provides complementary information on the quality of uncertainty attribution.

Our findings have important implications for \ac{xuq} research and practice. First, a multi-metric evaluation is necessary. Different metrics capture different quality aspects with varying reliability across datasets and methods. Second, we identified metric-specific limitations that provide actionable guidance for metric selection and future development. These include the high variability of feature flipping, the numerical instabilities of RIS, and the validity boundaries of UCS. Third, our framework provides a foundation for standardised evaluation, thereby facilitating reproducible comparisons of methods. Promising directions for future research include user studies evaluating whether functionally grounded metrics align with human judgments of uncertainty attribution utility, real-world applications in high-stakes domains, and investigations of the relationship between feature attribution quality and uncertainty attribution quality.